\documentclass{article}
\usepackage{ijcai26}

\usepackage{times}
\usepackage{soul}
\usepackage{url}
\usepackage[hidelinks]{hyperref}
\usepackage[utf8]{inputenc}
\usepackage[small]{caption}
\usepackage{graphicx}
\usepackage{amsmath}
\usepackage{amsthm}
\usepackage{booktabs}
\usepackage{algorithm}
\usepackage{algorithmic}
\usepackage[switch]{lineno}

\title{Meta-Judging with Large Language Models: Concepts, Methods, and Challenges}

\author{
Hugo Silva$^1$
\and
Mateus Mendes$^2$\And
Hugo Gonçalo Oliveira$^1$\\
\affiliations
$^1$University of Coimbra, CISUC/LASI – Centre for Informatics and Systems of the University of Coimbra, Department of Informatics Engineering\\
$^2$Polytechnic University of Coimbra, Rua da Misericórdia, Lagar dos Cortiços, S. Martinho do Bispo, 3045-093
Coimbra, Portugal.\\
RCM$^{2+}$ -- Research Centre in Asset Management and Systems Engineering\\
\emails
hugosilva@dei.uc.pt,
mmendes@isec.pt,
hroliv@dei.uc.pt
}

\begin{document}

\maketitle

\begin{abstract}
Large language models (LLMs) are evolving fast and are now frequently used as evaluators, in a process typically referred to as \textit{LLM-as-a-Judge}, which provides quality assessments of model outputs.
However, recent research points out significant vulnerabilities in such evaluation, including sensitivity to prompts, systematic biases, verbosity effects, and unreliable or hallucinated rationales.
These limitations motivated the development of a more robust paradigm, dubbed \textit{LLM-as-a-Meta-Judge}.
This survey reviews recent advances in meta-judging and organizes the literature, by introducing a framework along six key perspectives: (i)~Conceptual Foundations, (ii)~Mechanisms of Meta-Judging, (iii)~Alignment Training Methods, (iv)~Evaluation, (v)~Limitations and Failure Modes, and (vi)~Future Directions.
By analyzing the limitations of \textit{LLM-as-a-Judge} and summarizing recent advances in meta-judging by LLMs, we argue that \textit{LLM-as-a-Meta-Judge} offers a promising direction for more stable and trustworthy automated evaluation, while highlighting remaining challenges related to cost, prompt sensitivity, and shared model biases, which must be addressed to advance the next generation of LLM evaluation methodologies.
\end{abstract}

\section{Introduction}

Before the rise of ChatGPT in 2022, research on autonomous AI agents focused mainly on traditional frameworks like multi-agent systems (MAS) and expert systems \cite{lopez2009multi}, emphasizing social interaction and distributed intelligence. The concept of an AI agent, introduced by \cite{castelfranchi1998modelling}, has since evolved: modern AI agents now combine LLMs with external tools, function calling, and stepwise reasoning to perform complex, multi-stage tasks autonomously. By late 2023, Agentic AI emerged, where specialized agents collaborate to break down goals and coordinate efforts, letting multiple LLMs and intelligence counterparts leverage their unique strengths for more effective problem-solving \cite{chan2023chateval}.

Because human evaluators are limited and costly, the \textit{LLM-as-a-Judge} approach has gained traction, using LLMs to simulate human judgment \cite{chen2025multi}. Building on this, the concept of \textit{LLM-as-a-Meta-Judge} represents a promising step toward achieving more consistent and reliable automated evaluation. It underscores critical limitations that remain, namely high computational cost, strong sensitivity to prompt design, and the persistence of shared model biases. These factors continue to challenge alignment approaches like Direct Preference Optimization (DPO) \cite{rafailov2023direct}, highlighting the ongoing need for LLMs to serve as meta-judges that can mediate these shortcomings.

In essence, the meta-judge acts as a ``judge of judges'', ensuring evaluations are coherent, justified, and rubric-aligned. Despite growing interest, research in meta-judging remains sparse, and this work aims to define the concepts and unify disperse literature into a cohesive framework. The contributions of this paper can be summarized as follows:

\begin{enumerate}
\item At the definition level, we provide conceptual definitions of \textit{LLM-as-a-Judge} and \textit{LLM-as-a-Meta-Judge};
\item At the framework level, we conduct a systematic organization of fragmented literature, which is lacking from the literature, into a unified conceptual structure, namely the mechanisms of meta-judging, alignment training methods, evaluation of meta-judge, limitations and failure modes, and future directions.
\end{enumerate}

In Section \ref{sec:framework}, Figure \ref{fig:framework} summarizes the proposed framework of \textit{LLM-as-a-Meta-Judge} and the structure of the paper. Section 3 provides a review of survey research on \textit{LLM-as-a-Judge}. Section 4 presents the conceptual foundations of \textit{LLM-as-a-Judge} and \textit{LLM-as-a-Meta-Judge}. Section 5 describes the mechanisms of meta-judging, and Section 6 the alignment training methods. Section 7 is dedicated to evaluating the performance of the meta-judge, Section 8 discusses the limitations and failure methods, and Section 9 drafts the future directions for meta-judging. Finally, Section 10 presents the conclusions.

\section{Overview of the proposed LLM-as-a-Meta-Judge framework}
\label{sec:framework}

\begin{figure*}[!t]
    \centering
    \includegraphics[width=0.7\textwidth,height=0.2\textheight,keepaspectratio]{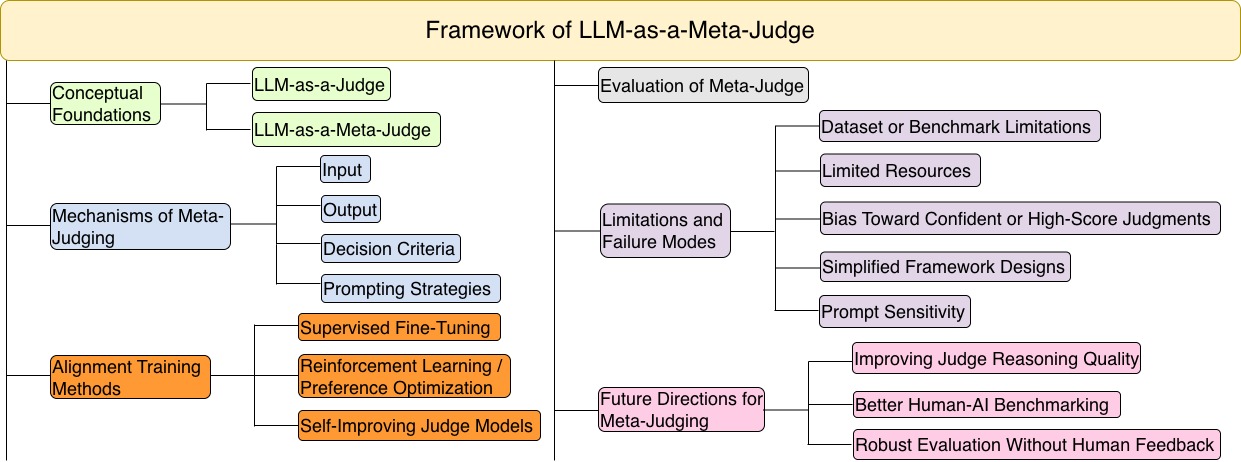}
    \caption{Proposed framework of LLM-as-a-Meta-Judge categorized in six main branches.}
    \label{fig:framework}
\end{figure*}

Figure~\ref{fig:framework} provides a structured overview of the proposed \textit{LLM-as-a-Meta-Judge} framework and serves as a conceptual map for the organization of this survey. Rather than presenting meta-judging as a single technique, the figure highlights it as a layered and evolving research paradigm that builds upon prior work on \textit{LLM-as-a-Meta-Judge}. The framework is organized into six interconnected branches, each corresponding to a major research dimension discussed in this paper.

The framework for LLM-based meta-judging begins with \textit{Conceptual Foundations}, which clarify the motivations, definitions, and assumptions behind judging and meta-judging. It then describes \textit{Mechanisms of Meta-Judging} and \textit{Alignment Training Methods}, detailing how models can assess other models’ judgments and be trained for consistency and reliability. \textit{Evaluation of Meta-Judge} covers metrics and protocols for assessing performance, while \textit{Limitations and Failure Modes} address challenges like bias and error propagation. Finally, \textit{Future Directions for Meta-Judging} highlights open research opportunities, illustrating a structured progression from foundational concepts to practical mechanisms, evaluation, and emerging questions.

\section{LLM-as-a-Judge}
\label{sec:llm_as_a_judge}

This section reviews recent surveys on LLMs as evaluators, outlining their definitions, development, validation, and practical use, while underscoring persistent challenges around bias, reliability, reproducibility, and ethics despite their scalability advantages.

The \textbf{Need for Automated Evaluation} is becoming increasingly clear. Traditional metrics such as BLEU \cite{papineni2002bleu} and ROUGE \cite{lin2004rouge} remain useful for surface-level assessment but often show weak correlation with human judgments in complex or creative tasks \cite{gu2025surveyllmasajudge,schluter2017limits}. As human evaluation is costly and difficult to scale, this gap has driven the adoption of the \textit{LLM-as-a-Judge} paradigm, where LLMs are used to evaluate generated outputs according to explicit criteria \cite{zheng2023judging,zhou2023lima}.

\textbf{LLM-as-a-Judge Paradigm and Taxonomies} explore the use of LLMs to evaluate, critique, or rank outputs rather than generate them, leveraging careful prompting, strong models, and structured evaluation pipelines to reduce human annotation effort \cite{gu2025surveyllmasajudge,zhuge2024agent,shinn2023reflexion}. These approaches support evaluation, training, and dataset construction through single models, ensembles, or human–AI collaboration, and are increasingly formalized through unified frameworks that define what to judge, how to judge, and how to assess outcomes \cite{li2024llms,li2025generation}.

\textbf{Biases, Robustness, and Improvement Strategies} are critical considerations for \textit{LLM-as-a-Judge} systems, which, while effective, suffer from recurring biases such as length, position, concreteness, and model-specific issues like egocentric bias and preference leakage \cite{li2025generation}. Existing mitigation approaches fall into prompt design, model-based enhancements, and output optimization, where techniques such as prompt engineering, fine-tuning, and aggregation of evaluation signals can improve reliability but may also introduce new biases or limit generalization \cite{gu2025surveyllmasajudge,hu2024llm}.

\textbf{Meta-Evaluation and Benchmarks} are central to assessing the reliability of \textit{LLM-as-a-Judge} systems, typically involving meta-evaluation against human judgments, alongside bias and robustness analyses. Prior and recent work relies on statistical agreement metrics (e.g., accuracy, correlation coefficients, Cohen’s kappa) and benchmark-based evaluations to measure alignment with human preferences, identify biases, and test performance under challenging or domain-specific conditions \cite{li2024llms,li2025generation}.

\textbf{Limitations and Open Challenges} in LLM-based evaluation include issues such as sensitivity to prompt changes, susceptibility to adversarial attacks, and reliance on specific model architectures, with research often focusing more on downstream tasks than on robust benchmarking \cite{gu2025surveyllmasajudge}. Ethical concerns stemming from biases in training data and opaque proprietary models have led to investigations into inference-time techniques like chain-of-thought (CoT) reasoning, self-consistency, debate-based judging, and human-in-the-loop approaches, though these solutions remain under-explored \cite{li2025generation}.

\section{Conceptual Foundations}

\label{subsec:conceptual_foundations}
This section introduces the conceptual foundations of automated evaluation with LLMs, focusing on two closely-related paradigms, \textit{LLM-as-a-Judge} and \textit{LLM-as-a-Meta-Judge}, as proposed in recent work on meta-judging. These concepts point to a transition away from fixed metrics and fully human-driven evaluation toward scalable, model-based assessment frameworks that rely on the reasoning capacity of LLMs.

\subsection{Concept of LLM-as-a-Judge}
\label{subsec:llm_as_a_judge}

The \textit{LLM-as-a-Judge} paradigm uses LLMs to evaluate outputs from humans or other models, offering flexible criteria and interpretable feedback, sometimes even replacing human evaluation \cite{li2025leveraging,li2024llms}. However, LLM judges can be biased, favor models similar to themselves, and occasionally hallucinate evaluation criteria.

Research has investigated a range of judging strategies, including prompting approaches such as zero-shot, few-shot, and CoT, as well as evaluation setups involving multiple judges or debate-based mechanisms. This work has also analyzed common biases, including verbosity, positional, and self-enhancement biases \cite{stahl2024exploring,zheng2023judging,verga2024replacing}. LLM judges also enable iterative self-improvement frameworks, like Self-Rewarding, though most focus on improving the evaluated outputs rather than the judge itself \cite{yuan2024self}.

Training LLM judges remains challenging: supervised fine-tuning (SFT) often lacks exposure to ambiguous or incorrect answers, limiting generalization, while methods like Self-Consistency and Best-of-N only partially address this \cite{wang2022self}. Preference-based approaches like DPO help models learn better from diverse outputs. DPO is a method used in training AI models to directly optimize for human preferences. Instead of just predicting what comes next in a text (like traditional language models), DPO tries to make the model align with what humans like or prefer. Overall, \textit{LLM-as-a-Judge} is promising for flexible, explainable evaluation, but single-agent setups are still vulnerable to bias and reliability issues \cite{ma2025judging,huang2023large,ye2024justice}.

\subsection{Concept of LLM-as-a-Meta-Judge}
\label{subsec:llm_meta_judge}

Research on using \textit{LLM-as-a-Judge} has advanced quickly, but a key issue remains: while LLMs are widely used to assess other systems, their own judgments are rarely scrutinized. Most studies focus on producing scores or feedback, with little attention to whether those evaluations are accurate, consistent, or biased. In effect, evaluation is automated without critically examining the evaluator itself, beyond comparing its output to human labels.

To address this, a new approach called \textit{LLM-as-a-Meta-Judge} has emerged. Here, LLMs don’t judge content directly. Instead, they evaluate other judgments, namely the scores, rationales, or critiques produced by LLMs or humans, acting as ``judges of judges''. Meta-judges can detect flawed reasoning, identify systematic bias, and even revise judgments. Table \ref{tab:llm_judge_vs_llm_meta_judge} summarizes the differences between traditional LLM judges and meta-judges. Although promising, this approach is still in its early stages.

\begin{table*}[t]
    \centering
    \footnotesize
    \begin{tabular}{|p{3cm}|p{5cm}|p{7cm}|}
        \hline 
        \textbf{Aspect} & \textbf{LLM-as-a-Judge} & \textbf{LLM-as-a-Meta-Judge} \\ \hline
        What is evaluated & Model outputs & Judge evaluation of model outputs \\ \hline
        Goal & Produce a score, ranking, or judge & Validate, correct, or critique the judge's evaluation \\ \hline
        Input & Prompt and model answers & Prompt, model answers, and judge's evaluation \\ \hline
        Role & Primary evaluator & Secondary evaluator or auditor \\ \hline
        Tries to address & Need for automated judgments & Bias, inconsistency, or error in LLM judges \\ \hline
    \end{tabular}
    \caption{Key differences between LLM-as-a-Judge and LLM-as-a-Meta-Judge.}
    \label{tab:llm_judge_vs_llm_meta_judge}
\end{table*}

Compared to humans, LLM-based meta-judging is faster and more scalable. However, prior work has focused on aligning LLM judgments with human preferences, often ignoring the biases present in human evaluations. Selecting reliable judgments from multiple LLM evaluations is another underexplored challenge. To tackle this, \cite{li2025leveraging} propose a three-stage pipeline to filter and audit judgments, while \cite{wu2025meta} introduce a Meta-Rewarding stage where a model can iteratively improve its own evaluations during training. Other methods, like \cite{trivedi2024self}, show that using a judge’s own rationales via Self-Rationalization and DPO can significantly improve accuracy. Bias in meta-judging remains an open question, where \cite{ma2025judging} provide one of the first systematic analyses, examining position, verbosity, chain-of-thought, and bandwagon effects.

Several meta-judging implementations exist. \cite{li2025leveraging} use a multi-agent setup, where multiple LLMs independently or collaboratively score judgments, and then aggregate results to estimate reliability. \cite{wu2025meta} propose an unsupervised, iterative pipeline where a single model alternates between acting as actor, judge, and meta-judge, creating preference pairs to train judges via DPO. \cite{trivedi2024self} generate multiple judgments with rationales for the same input, refining reasoning-to-score alignment. \cite{ma2025judging} explore two meta-judging variants: selecting the best judgment from a pool, or synthesizing a final judgment from multiple inputs, evaluating the ability to integrate and reconcile diverse information.

Empirical results are encouraging. \cite{li2025leveraging} report over 15\% precision gains on JudgeBench and 8\% improvement over single-agent baselines. \cite{wu2025meta} show win-rate improvements from 22.9\% to 39.4\% on AlpacaEval 2 and 20.6\% to 29.1\% on Arena-Hard using Meta-Rewarding. \cite{trivedi2024self} find Self-Rationalization improves rationales, with a 62\% win rate, outperforming even larger models trained with other methods. \cite{ma2025judging} find that meta-judging achieves bias consistency comparable to single-model judging, with modest improvements in position bias.

Recent work has explored a range of models for evaluation and judgment tasks. For example, \cite{li2025leveraging} benchmarked several advanced LLMs, including GPT-4o, Claude-3.5-Sonnet, and LLaMa-3.1-405B-Instruct, while \cite{wu2025meta} and \cite{yuan2024self} investigated instruction-finetuned Llama-based judges and their correlation with GPT-4-1106-preview. Other studies trained specialized Llama-3.1 judge models for pairwise and point-wise evaluation and compared them with widely used open-source judges such as Prometheus and Auto-J \cite{trivedi2024self}. Additionally, meta-judge configurations combining models like GPT-4o-mini, LLaMA-3.3-70B, and DeepSeek-V3 have been examined, with some models reserved for reference judgments rather than meta-evaluation \cite{ma2025judging}.

Overall, while using LLMs as meta-judges is still an emerging area, current results indicate it is a promising direction for building more reliable, transparent, and self-aware evaluation systems.

\section{Mechanisms of Meta-Judging}
\label{section:mechanisms_meta_judging}

The mechanisms of meta-judging can be categorized into four components: the input structure for the meta-judge, output, decision criteria, and prompting strategies.

\subsection{Input Structure for Meta-Judge}
\label{subsec:input_structure_meta_judge}

These works explore how meta-judging can be improved by giving judges richer and more transparent inputs, rather than limiting them to just the original prompt and a single candidate answer. Across studies, judges are provided with additional context such as model rationales or critiques \cite{li2025leveraging}, the actor model’s original response and prompt, pairs of judgments with the largest score variance, and the evaluation rubric itself \cite{wu2025meta}. Other approaches include supplying multiple judgments, each with an explicit score and rationale \cite{trivedi2024self}, or combining responses with responses to the prompt, earlier referees’ judgments, scores and critiques \cite{ma2025judging}.

Overall, these methods emphasize evaluating the reasoning process behind an answer, not just the final output. As noted by \cite{li2025leveraging}, this shift toward multi-component and transparent inputs helps judges better understand how conclusions are reached.

Some works also focus on reducing bias in the judging process. For example, \cite{wu2025meta} address positional bias by asking the model to judge the same pair of responses twice with their order swapped, and by applying weighted scoring depending on whether a response appears first or second.

Finally, \cite{trivedi2024self} describe a Self-Rationalization approach in which the same model generates multiple independent judgments for a single input, each with its own score and rationale. This allows for more diverse perspectives on the same response and can lead to more robust evaluations.

\subsection{Output}
\label{subsec:output}

The outputs of meta-judge are the meta-judge score used to filter unreliable judgments by retaining only those with high-quality, well-explained, and decisive conclusions \cite{li2025leveraging}, generation of preference pairs of judgments, with CoT reasoning, followed by its choice of better and worst judgment, to train the judge model via DPO \cite{wu2025meta,trivedi2024self}, best judgment from a pool as the final verdict, or generation of its own judgment based on the pool to serve as the final verdict \cite{ma2025judging}. \cite{li2025leveraging} refer that by defining a threshold, judgments scoring above this value are considered reliable and correct, while those below are discarded. This enables precision-focused evaluation by minimizing false positives, ensuring that incorrect meta-judgments are classified as correct. \cite{wu2025meta} mention that their approach allows to compute scores that account for the positional bias in the meta-judge evaluations, providing a more accurate reward signal representing the judgment quality. \cite{trivedi2024self} claim that the preference pairs are used to finetune the model via DPO resulting in an improvement of the rationale generation and response evaluation.

\subsection{Decision Criteria}
\label{subsec:decision_criteria}

Recent work on meta-judging shows that there is no single standard for how evaluation decisions should be made. Instead, different studies adopt different assumptions about what makes a judgment reliable and how to reduce bias and uncertainty when combining multiple evaluations.

\cite{li2025leveraging} propose a multi-agent setup where several agents act as meta-judges. Agents can score responses independently or interact, and the final score is determined via one of three strategies: weighted averaging (scores are combined based on criterion importance and agent reliability), majority voting (responses passing a threshold from most agents get a high score), or panel discussion (agents assume roles, discuss, and reach consensus, sometimes summarized by a dedicated agent). This discussion-based approach helps reduce bias and improve score quality.

\cite{wu2025meta} take a different route, avoiding scalar scores. They focus on pairwise comparison of judgments, especially those with high uncertainty. Each comparison involves a meta-judge evaluating which judgment is better, considering the prompt, response, and evaluation rubric. To reduce positional bias, the order of judgments is swapped and weighted, and pairwise results are converted into continuous meta-rewards via Elo scores and maximum likelihood estimation \cite{zheng2023judging}. They also identify and correct for a length bias that favors longer judgments.

\cite{trivedi2024self} emphasize generating high-quality preference data through ``Self-Rationalization''. They create preference pairs by linking a chosen judgment with a rejected one, using: (1) ground-truth correctness, optionally applying a margin to emphasize differences; (2) a meta-judge model \cite{wu2025meta} to rank judgments based on correctness and reasoning quality; or (3) majority voting/self-consistency, selecting the most common score among multiple judgments. These methods aim to produce cleaner, more informative training signals.

\cite{ma2025judging} investigate the fundamental structure of meta-judging. They avoid techniques like rubric engineering \cite{li2025leveraging} to isolate bias effects, control for positional bias by shuffling outputs, and vary the number of judges to study how judge pool size influences observed biases.

\subsection{Prompting Strategies}
\label{subsec:prompting_strategies}

Research has explored a range of meta-judge frameworks that vary in how judgments are compared, scored, and synthesized, spanning single-agent prompting strategies, rubric-based comparisons, and multi-judgment aggregation.

Prior work has explored meta-judging mainly in single-agent settings. \cite{li2025leveraging} identify four prompt strategies for this scenario. The baseline meta-judge \cite{wu2025meta} produces a single overall score without breaking it down by criteria such as accuracy or instruction adherence. The short scoring rubric introduces a concise set of criteria to generate a soft label score, while the long scoring rubric expands this idea with detailed, multi-sentence descriptions for each criterion and score range. Finally, the binary rubric asks the judge to decide whether all criteria are met using a simple True/False judgment. \cite{wu2025meta} propose a meta-judge prompt for comparing two existing judgments of the same model response. The evaluator reviews the original question, the response, and both judgments, then decides which judgment is better. This decision is justified by checking how well each judgment follows the rubric, how accurate it is, and how consistent its reasoning is, ending with a clear statement of the preferred judgment. \cite{trivedi2024self} describe that in the \textit{LLM-as-a-Meta-Judge} self-assessment setting, the model evaluates the quality of its own judgments, rating them based on factors such as reasoning quality and scoring accuracy. \cite{ma2025judging} introduce two meta-judging settings. In the ``Select'' setting, the meta-judge looks over multiple judgments, summarizes their key points, notes where it agrees or disagrees, and then picks the strongest option. In the ``Conclude'' setting, it combines all the judgments into a single final decision, scores assistants on dimensions like helpfulness and accuracy, and reflects on the referees’ judgments. Together, these settings test how well a meta-judge can reason over, integrate, and reconcile multiple perspectives.

\section{Alignment Training Methods}
\label{section:alignment_training_methods}

\textbf{Supervised Fine-Tuning} serves as the foundational alignment stage in meta-judge training, equipping the base model with initial evaluative capabilities by learning to produce human-aligned judgments from labeled evaluation data.

\cite{wu2025meta} describe a preparation process for Meta-Rewarding training. First, the seed model (Llama-3-8B-Instruct) undergoes supervised fine-tuning on the Evaluation Fine-Tuning (EFT) dataset from \cite{yuan2024self}, which is built from Open Assistant \cite{kopf2023openassistant}. This dataset contains ranked human responses and is designed to train the model as an \textit{LLM-as-a-Judge}, providing the evaluative capability needed before applying Meta-Rewarding.

\cite{trivedi2024self} show that Self-Rationalizing Evaluators (SREs) perform better than standard SFT-trained \textit{LLM-as-a-Judge} models, both in the coherence of their rationales and in scoring accuracy. They point out a core weakness of standard SFT: it mainly teaches the model to recognize correct judgments, without explicitly exposing it to incorrect or borderline cases. As a result, such judges often struggle with ambiguous or non-binary evaluations. In their setup, a base judge model is first trained via SFT on labeled pointwise (e.g., Likert-scale) and pairwise evaluation data, and then further calibrated in later stages to improve the consistency and reliability of its judgments.

\textbf{Reinforcement Learning/Preference Optimization} methods train models using preference data, often produced by judge models, to better align outputs with desired behaviors. While \cite{li2025leveraging} do not directly propose DPO or Reinforcement Learning from Human Feedback (RLHF), their analysis of failure modes in LLM-based judges sheds light on the types of errors these approaches seek to address and how improved preference datasets can be constructed. RLHF explicitly optimizes for human-aligned objectives such as helpfulness, safety, and overall quality, making it more complex than DPO, which instead learns directly from preference pairs without an explicit reinforcement learning loop. \cite{wu2025meta} propose Meta-Rewarding, an iterative framework where a single LLM cycles through the roles of actor, judge, and meta-judge to progressively improve evaluation quality. The actor generates multiple responses, the judge ranks them to form preference pairs, and the meta-judge compares alternative judgments to produce training data that refines the judge, optimized using DPO for stability and simplicity. The method explicitly mitigates length bias by favoring shorter outputs when scores are similar \cite{singhal2023long}. Experiments show that Meta-Rewarding significantly outperforms Self-Rewarding in agreement with GPT-4 and human evaluators, although improvements slow down in later iterations, likely due to distribution shifts between model-generated and human-written responses. \cite{trivedi2024self} propose combining SFT with DPO to improve judging performance. SFT first establishes basic task competence, after which DPO refines the model using preference pairs that distinguish better judgments from worse ones, allowing the model to learn more effectively from negative examples. These preference pairs are built using three strategies: pairing correct judgments against incorrect ones based on ground truth, meta-judge ranking following \cite{wu2025meta}, and majority voting across multiple judgments. Results show that using SFT and DPO together outperforms either method alone. While SFT teaches general task behavior, DPO better aligns the model with preference-based reasoning and avoids the diluted training signals that can arise from long rationales in SFT or RLHF.

\textbf{Self-Improving Judge Models} represent approaches in which LLMs iteratively refine their own evaluations, enhancing judgment quality and reducing biases without relying on external human supervision. Recent research has explored how LLMs can act as judges and even improve their own evaluation abilities. \cite{li2025leveraging} point out that most existing work tries to align LLM judgments with human preferences, but often ignores the fact that human judgments themselves can be biased or inconsistent. Earlier studies have also identified several weaknesses in \textit{LLM-as-a-Judge} systems, including positional and verbosity biases, self-enhancement, and shallow reasoning, especially for open-ended evaluations \cite{zheng2023judging}. To address these issues, recent work has proposed letting models reflect on and revise their own judgments through meta-judging, reducing reliance on human supervision. Along these lines, \cite{wu2025meta} propose Meta-Rewarding, a method that allows a model to assess and refine its own judgments using internal feedback. Their results suggest that models can meaningfully improve their evaluation skills autonomously. Similarly, \cite{trivedi2024self} introduce SRE, where a model generates, compares, and learns from its own evaluations, using DPO to iteratively improve reasoning quality and scoring accuracy without additional human input. Finally, \cite{ma2025judging} show that using LLMs as meta-judges improves consistency and can reduce intrinsic biases, particularly when evaluating larger judgment pools.

\section{Evaluation of Meta-Judge}
\label{section:evaluation_meta_judge}

This section reviews recent empirical efforts to evaluate meta-judging frameworks, focusing on their effectiveness across different tasks, datasets and benchmarks, judging strategies, and agent configurations. 

\cite{li2025leveraging} propose evaluating LLM judgments beyond accuracy by considering reasoning quality, interpretability, and error detection. They use JudgeBench \cite{tan2024judgebench}, which integrates MMLU-Pro, LiveBench, and LiveCodeBench \cite{wang2024mmlu,white2024livebench,jain2024livecodebench}, covering knowledge, reasoning, math, and coding tasks with responses from GPT-4o and Claude-3.5-Sonnet annotated by humans. Benchmarking GPT-4o, GPT-4o-mini, Claude-3.5-Sonnet, and LLaMa-3.1-405B-Instruct, they find that judging strategies should match task type: task-aligned models help for knowledge tasks, stronger models and longer rubrics aid complex reasoning, and detailed rubrics benefit math. Multi-agent approaches, like majority voting, weighted averaging, and panel discussions, further improve precision depending on the task. \cite{wu2025meta} introduce Meta-Rewarding, a self-play framework that improves a model’s ability both to answer questions and to evaluate responses. Building on prior self-rewarding work \cite{yuan2024self}, they fine-tune Llama-3-8B-Instruct using AlpacaEval 2, Arena-Hard, and MT-Bench. The model alternates between three roles (actor, judge, and meta-judge), generating its own preference data while explicitly controlling for length and positional bias. Actor training uses scored responses with length-controlled preference selection, while judge training focuses on cases with high uncertainty and relies on pairwise meta-judging with Elo ranking. Meta-Rewarding substantially outperforms Self-Rewarding and Self-Play Preference Optimization (SPPO) baselines, raising win rates from 22.9\% to 39.4\% and approaching the performance of Claude Opus. It improves performance across most instruction categories, strengthens judging quality, and maintains multi-turn ability, though alignment gains with human judgments diminish in later iterations due to distribution shift. \cite{trivedi2024self} explore SREs, which learn to judge responses with explicit reasoning but without human-annotated preference data. They evaluate on RewardBench, BiGGen Bench, and Feedback Bench. Starting from a base judge trained via SFT, they iteratively generate multiple judgments with rationales and construct preference pairs using three strategies: matching ground truth scores, meta-judge selection \cite{wu2025meta}, and majority voting. Training with DPO on these preferences leads to evaluators that outperform the seed model and stronger baselines while using fewer samples and less computation. The results show that explicit rationales help judging performance when paired with high-quality preference data, whereas long rationales trained via SFT or RLHF can add noise. Higher-margin preference pairs consistently lead to better alignment and evaluation quality. \cite{ma2025judging} investigate whether meta-judge frameworks can reduce bias in LLM evaluation. They train meta-judges on pairwise comparisons from multiple LLM judges using MT-Bench and CALM-Alignment. Results show that diverse judge pools can match the bias consistency of strong single judges, with larger pools notably improving position-bias consistency. Overall consistency may slightly decrease, and adding a bias-free agent has limited effect unless the meta-judge can explicitly identify and weight it.

As summarized in Table \ref{tab:main_results_meta_judging}, different meta-judging frameworks emphasize distinct goals, ranging from bias reduction to autonomous judge improvement.

\begin{table*}[t]
    \centering
    \footnotesize
    \begin{tabular}{|p{2cm}|p{3cm}|p{12cm}|}
        \hline 
        \textbf{Work} & \textbf{Core Contribution} & \textbf{Main Results / Findings} \\ \hline
        \cite{li2025leveraging} & Analysis of failure modes and design choices & Providing judges with clear rationales or rubrics increases reliability, precision-oriented filtering lowers false positives, and multi-agent judging consistently outperforms single evaluators, though the best aggregation method depends on the task.\\ \hline
        \cite{wu2025meta} & Meta-Rewarding framework for self-improving judges & Iterative meta-judging with bias-controlled pairwise comparisons boosts judging quality and actor performance, nearing strong proprietary model results, though gains plateau over iterations.\\ \hline
        \cite{trivedi2024self} & Self-Rationalizing Evaluators &  Training with DPO on multiple rationalized judgments improves scoring accuracy and reasoning quality over standard SFT, with larger-margin preference pairs boosting alignment, and combining SFT with DPO outperforming each individually.\\ \hline
        \cite{ma2025judging} & Structural study of meta-judging and bias & Meta-judging can reduce positional bias and improve consistency with larger, diverse judge pools, though overall consistency gains are not guaranteed and adding ``bias-free'' agents alone has limited effect.

        \\ \hline
        
    \end{tabular}
    \caption{Summary of main results in meta-judging research.}
    \label{tab:main_results_meta_judging}
\end{table*}

\section{Limitations and Failure Modes}
\label{section:limitations_failure_modes}

\textbf{Dataset or Benchmark Limitations} play a central role in shaping the conclusions of meta-judge studies, as the scope and composition of the datasets can constrain generalization and affect evaluation outcomes. Several limitations and open questions emerge from prior work. \cite{li2025leveraging} evaluate a meta-judge selection framework using a relatively small set of 350 judgments based on response pairs generated by GPT-4o-mini. This limited judgment pool may restrict how well their findings generalize, as the framework itself constrains the diversity and scale of evaluations. \cite{trivedi2024self} point out a gap in understanding how self-curated datasets, especially those expanded through sampling techniques, affect the quality of rationales and scoring in \textit{LLM-as-a-Judge} settings. Their observations suggest opportunities to study whether combining DPO with more advanced sampling strategies can lead to better rationales and stronger overall performance. Similarly, \cite{ma2025judging} report that their experiments were conducted on only two benchmarks, which may limit the robustness of their conclusions. While additional benchmarks such as Judge-Bench \cite{tan2024judgebench}, which focuses on problem-solving ability, are available, they were intentionally excluded to avoid confounding bias measurements with task-solving errors. Finally, limited computational resources remain a major challenge. As noted by \cite{ma2025judging}, multi-agent inference is expensive and heavily dependent on strong instruction-following, which restricts evaluation scale. This makes fine-grained statistical analysis difficult leading to higher variance and less reliable results.

\textbf{Bias Toward Confident or High-Score Judgments} is a key factor influencing meta-judge performance, as judges often favor responses they perceive as more confident or assign higher scores more readily. \cite{wu2025meta} identify several limitations in their experimental setup from using a 5-point judging system. The scoring scale led to frequent ties and score saturation, making it difficult to distinguish small quality differences or detect improvements as training progressed. Despite attempts to mitigate positional bias in the meta-judge, the bias persisted and avoided further gains by the third iteration. Additionally, the judge showed a tendency to over-assign high scores, further reducing the ability to discriminate between responses, and exhibited limited improvement when evaluating responses not generated by itself. \cite{ma2025judging}, regarding how biases manifest in multi-agent frameworks when used in the context of \textit{LLM-as-a-Judge} systems, highlight that previous work is limited by examining only a small set of biases, position, bandwagon, CoT, and verbosity, which fails to capture the full range of possible biases. They also note that bias mitigation is underexplored, as only a single method for reducing position bias is evaluated, leaving effective bias mitigation an open and ongoing challenge.

\textbf{Simplified Framework Designs} are central to this discussion, as \cite{ma2025judging} points out the applied simplified multi-agent framework, in order to maintain clarity and control over confounding variables, limiting the direct applicability of the results to more advanced multi-agent architectures, but this choice allows to isolate bias-related effects without interference from more elaborate coordination strategies.

\textbf{Prompt sensitivity} is a key factor, as \cite{ma2025judging} discusses how the structure of prompts given to automated judge models can itself limit evaluation outcomes.

\section{Future Directions for Meta-Judging}
\label{section:future_directions_meta_judging}

\textbf{Improving Judge Reasoning Quality} is a central challenge in meta-judging research. Recent work shows that stronger evaluation depends on a judge’s ability to generate and distinguish high- versus low-quality reasoning, supported by high-quality preference data and training strategies such as DPO combined with advanced sampling \cite{trivedi2024self}. Other studies reveal that judgment errors often stem from systematic biases, such as position, bandwagon, CoT, and verbosity effects, and argue that mitigating these issues will require better prompt design, alternative judge architectures, and exploration of additional bias sources \cite{ma2025judging}.

Regarding \textbf{Better Human-AI Benchmarking}, recent work points to the need for more robust and nuanced approaches to evaluating human–AI systems, both by leveraging strong LLMs as meta-judges to produce high-quality preference data and by designing benchmarks that disentangle evaluator bias from underlying model capabilities. \cite{li2025leveraging} argue that the strong performance of LLMs as meta-judges can be leveraged to generate high-quality preference datasets. \cite{ma2025judging} highlight the need for more benchmarks to distinguish bias effects from capability failures. They suggest creating controlled, bias-only environments to separate evaluator bias from task performance and propose scalable strategies for multi-agent LLM evaluation dues to computational constraints.

\textbf{Robust Evaluation Without Human Feedback} represents a promising future direction for meta-judging, focusing on the development of autonomous and scalable evaluation frameworks that minimize or eliminate reliance on human supervision. \cite{li2025leveraging} highlight the potential of using LLMs to train other LLMs as evaluators, aiming at fully autonomous, scalable, and accurate assessment systems without human supervision. \cite{wu2025meta} highlight that models improving them selves without human feedback is a promising approach to address the super alignment challenge \cite{burns2023weak}, aiming for models that can potentially surpass human-level judgment abilities.

\section{Conclusion}
\label{sec:conclusion}

The evolution from \textit{LLM-as-a-Judge} to \textit{LLM-as-a-Meta-Judge} marks a shift from using LLMs to make judgments to using them to evaluate such judgments. While early LLM judges offered scalable, flexible alternatives to humans, they struggled with bias, inconsistency, and unquestioned authority. Meta-judges address this by scrutinizing judgments for correctness, reasoning, and fairness, adding a layer of oversight. However, they come with higher costs, sensitivity to prompts, and imperfect bias detection, so they should be seen as a complement, not a replacement, to careful evaluation.

To support this shift, the paper offers a comprehensive survey of meta-judging, an area that has received little systematic attention so far. It first clarifies the concepts of \textit{LLM-as-a-Judge} and \textit{LLM-as-a-Meta-Judge}, introducing a contextualized definition that considers the evaluation target, goals, inputs, and model properties. Second, it then organizes prior work into a coherent framework along the key dimensions of meta-judging, as summarized in Figure~\ref{fig:framework} of Section \ref{sec:framework}. This framework structures the literature around conceptual foundations, meta-judging mechanisms, alignment training methods, evaluation protocols, and known limitations, bringing together previously fragmented research while highlighting open problems and promising directions for future work.

Using multiple layers of LLMs to judge each other has limited value, since these models share similar biases, and meta-judging mostly checks internal consistency rather than truth. Human judgment remains essential, especially for subtle or value-laden tasks. Ultimately, understanding and refining how we evaluate AI is the key to building systems we can trust.

\section*{Acknowledgments}

This work is financed through national funds by FCT - Fundação para a Ciência e a Tecnologia, I.P., in the framework of the Project UIDB/00326/2025 and UIDP/00326/2025.

\bibliographystyle{named}
\bibliography{ijcai26}

\end{document}